\documentclass[conference]{IEEEtran}
\IEEEoverridecommandlockouts
\usepackage{cite}
\usepackage{amsmath,amssymb,amsfonts}
\usepackage{algorithmic}
\usepackage{graphicx}
\usepackage{textcomp}
\usepackage{xcolor}
\usepackage{booktabs}
\usepackage{tabularx}
\usepackage{multirow}
\usepackage{hyperref}
\usepackage{bm}

\newcommand{\tens}[1]{\bm{#1}}
\newcommand{\vect}[1]{\bm{#1}}
\newcommand{\nq}{N_{\mathrm{q}}}
\newcommand{\ncam}{N_{\mathrm{cam}}}
\newcommand{\hh}{H_{\mathrm{h}}}
\newcommand{\cl}{C_{\mathrm{L}}}
\newcommand{\hp}{H_{\mathrm{p}}}
\newcommand{\wpp}{W_{\mathrm{p}}}
\newcommand{\tfuture}{T_{\mathrm{future}}}

\DeclareMathOperator{\Sample}{Sample}
\DeclareMathOperator{\GridSample}{GridSample}
\DeclareMathOperator{\BEVSample}{BEVSample}
\DeclareMathOperator{\maskedsoftmax}{masked\_softmax}
\DeclareMathOperator{\invSigmoid}{inverse\_sigmoid}

\def\BibTeX{{\rm B\kern-.05em{\sc i\kern-.025em b}\kern-.08em
    T\kern-.1667em\lower.7ex\hbox{E}\kern-.125emX}}
\begin{document}

\title{Li-ViP3D++: Query-Gated Deformable Camera–LiDAR Fusion for End-to-End Perception and Trajectory Prediction\\

\thanks{Published in IEEE Access. The version of record is available at DOI: 10.1109/ACCESS.2026.3709080. This work is licensed under CC BY 4.0. \\This work was funded by the Excellence Strategy of the German Federal and State Governments and the Helmholtz Program “Engineering Digital Futures”. Additionally, it was funded thanks to the support of the Program Slovakia for the project: Research, development, testing, and validation of the integration of AI-driven 5G/6G with edge computing for ultra-low-latency, reliable, and secure communication systems with potential applications in industry, transportation, and healthcare (401101C689), co-financed by the European Union. This publication was also supported by the Slovak Research and Development Agency under Contract No. APVV-23-0519. Finally, the support of the ELLIIT strategic research network and the Helmholtz Association Initiative and Networking Fund on the HAICORE@KIT partition is gratefully acknowledged.}
}

\author{\IEEEauthorblockN{
Matej Halinkovic\IEEEauthorrefmark{1},
Nina Masarykova\IEEEauthorrefmark{1},
Alexey Vinel\IEEEauthorrefmark{2}\IEEEauthorrefmark{3},
Marek Galinski\IEEEauthorrefmark{1}}
\IEEEauthorblockA{\IEEEauthorrefmark{2}
Karlsruhe Institute of Technology, Kaiserstraße 89, 76133 Karlsruhe, Germany}
\IEEEauthorblockA{\IEEEauthorrefmark{1}
Slovak University of Technology, Ilkovičova 2, 842 16 Bratislava, Slovakia}

\IEEEauthorblockA{\IEEEauthorrefmark{3}
Halmstad University, Kristian IV:s väg 3, 301 18 Halmstad, Sweden}
{Email: \{matej.halinkovic, nina.masarykova, marek.galinski\}@stuba.sk}, alexey.vinel@kit.edu}

\maketitle

% TODO rewrite
\begin{abstract}
% Problem: end-to-end perception+prediction is promising but often depends on HD maps and heavy vision backbones; this is brittle for deployment.

% Core idea: map-free Li-ViP3D++ variant with a new agent interaction module (non-VectorNet), and multimodal agent queries that permit lower-res RGB while maintaining performance.

% Results: (i) removing HD map + replacing interaction improves metrics vs baseline, (ii) low-res RGB keeps performance, (iii) ~75 ms inference.

% Close with VNC relevance: practical real-time PnP for connected mobility / safety.

End-to-end perception and trajectory prediction from raw sensor data is one of the key capabilities for autonomous driving. Modular pipelines restrict information flow and can amplify upstream errors. Recent query-based, fully differentiable perception-and-prediction (PnP) models mitigate these issues, yet the complementarity of cameras and LiDAR in the query-space has not been sufficiently explored. Models often rely on fusion schemes that introduce heuristic alignment and discrete selection steps which prevent full utilization of available information and can introduce unwanted bias. We propose Li-ViP3D++, a query-based multimodal PnP framework that introduces Query-Gated Deformable Fusion (QGDF) to integrate multi-view RGB and LiDAR in query space. QGDF (i) aggregates image evidence via masked attention across cameras and feature levels, (ii) extracts LiDAR context through fully differentiable BEV sampling with learned per-query offsets, and (iii) applies query-conditioned gating to adaptively weight visual and geometric cues per agent. The resulting architecture jointly optimizes detection, tracking, and multi-hypothesis trajectory forecasting in a single end-to-end model. On nuScenes, Li-ViP3D++ improves end-to-end behavior and detection quality, achieving higher EPA (0.505) and mAP (0.616) while substantially reducing false positives (FP ratio 0.069), and it is faster than the prior Li-ViP3D variant (139.82 ms vs. 145.91 ms). Additional experiments were performed to evaluate the impact of the reduced resolution of RGB inputs and missing HD maps on the behavior of the model. The results of the experiments indicate that query-space, fully differentiable camera–LiDAR fusion can increase the robustness of end-to-end PnP without sacrificing deployability.
%Overall, our results suggest that agent-query PnP architectures can deliver stronger prediction with fewer external dependencies, supporting scalable, real-time safety applications in vehicular networks.

\end{abstract}


\section{Introduction}
Forecasting the positions and future behavior of agents in a traffic scene is one of the key capabilities needed for autonomous driving, for road-infrastructure management, and ensuring the safety of vulnerable road users. For downstream planning and control, anticipating how surrounding participants will evolve is as important as estimating their current state. As trajectory prediction methods improve in autonomous driving environments, they contribute directly to safer interactions among road users and to smoother traffic flow.

Among Perception and Prediction (PnP) approaches in autonomous driving, the most advanced direction is to predict trajectories directly from raw sensor inputs. Accordingly, end-to-end models that unify object detection, tracking, and prediction have become an important component of embodied perception~\cite{peri2022forecasting_lidar_multiple}. However, leading solutions still omit several practical considerations. In particular, there remains a lack of methods that can fully exploit both RGB and LiDAR streams in a single, fully differentiable framework. Although unimodal methods have historically dominated computer vision for dynamic scenes, they exhibit fundamental shortcomings~\cite{bijelic2020unimodal_bad_1, geiger2012unimodal_bad_2}. LiDAR-centric approaches often emphasize computational efficiency, whereas camera-based systems can be deployed using comparatively inexpensive hardware. Despite these advantages, unimodal pipelines inherit the failure modes and constraints of the single sensor they depend on.

These limitations reduce the reliability of PnP systems in challenging conditions that involve diverse object types and changing environments. Recent research has shown that object queries proposed by DETR3D~\cite{carion2020detr,wang2022detr3d}, a learned sparse set of embedding vectors (object priors) which can be decoded into a 3D reference point, are an effective interface for using RGB data for 3D detections. Query-based approaches have expanded to showcase the suitability of the interface for object tracking and trajectory prediction~\cite{gu2023vip3d}. However, while they underline the value of RGB for 3D detection and forecasting, they do not resolve the broader uncertainty issues faced by PnP systems that rely exclusively on vision. This is especially relevant when the goal is to protect vulnerable road users. Recent research on traffic monitoring with social robots~\cite{ari1,ari2} indicates that RGB sensing can support basic detection and intention inference, yet it does not provide sufficiently accurate estimates of agent position and timing.

%While query-based approaches built around agent queries~\cite{carion2020detr,wang2022detr3d,gu2023vip3d} underline the value of RGB for 3D detection and forecasting, they do not resolve the broader uncertainty issues faced by PnP systems that rely exclusively on vision. This is especially relevant when the goal is to protect vulnerable road users. Recent research on traffic monitoring with social robots~\cite{ari1,ari2} indicates that RGB sensing can support basic detection and intention inference, yet it does not provide sufficiently accurate estimates of agent position and timing.

This manuscript extends our earlier proceedings publication~\cite{halinkovic2025LiViP3D}. In that work, we introduced Li-ViP3D, a novel end-to-end and fully differentiable approach to trajectory prediction from combined RGB and LiDAR data. Li-ViP3D expanded the agent-query paradigm into a multimodal framework capable of predicting trajectories over a 6-second horizon for a broad set of agents, including cars, bicycles, and pedestrians. We also redesigned the 3D agent-query interface to incorporate LiDAR information and showed that this interface enables effective fusion of RGB and LiDAR cues.

In the present extension, our primary contributions focus on a new query-gated deformable fusion (QGDF) strategy that further increases the practicality of agent queries for end-to-end perception and prediction. We modify our Li-ViP3D architecture to use the QGDF mechanism. The resulting Li-ViP3D++ network improves computational efficiency and results in terms of both prediction and detection quality, while also enhancing the explainability of the base Li-ViP3D. The proposed architecture is evaluated against baselines under various conditions, demonstrating the robustness of the query-based fusion method against reduced RGB resolution and missing HD maps. Li-ViP3D++ source code is publicly available\footnote{https://github.com/mathali/Li-ViP3D-PP}.

% Set up the deployment friction:
% HD maps are a major operational dependency (creation, updates, distribution).

% RGB quality and compute budget limit edge deployment.
% One paragraph positioning: “We show the agent-query multimodal interface can reduce reliance on both HD maps and high-res RGB.”

% Bullet contributions (3 bullets, very concrete):

% Map-free variant + new interaction module

% Ablations showing performance improves without HD map and with the new interaction module

% Multimodal queries enable lower-res RGB and 75 ms runtime without losing accuracy

\section{Related Work}

\subsection{Traditional Perception and Prediction Pipelines}
Research on the detection-tracking-forecasting pipeline is extensive and spans many facets of these challenging problems~\cite{huang2022surveytp}. Although numerous methods address parts of the pipeline, the shift toward fully end-to-end formulations is still in its early stages. As a result, there is not yet broad agreement on how to design end-to-end systems that simultaneously account for the full range of requirements.

Conventional PnP systems typically decompose the task into separate components~\cite{xu2024valeo4cast,wang2023modular1,woo2023modular2,xu2024towardsmodular}: object detection, object tracking, and trajectory prediction. Advocates of modular designs emphasize their adaptability, ease of upgrading individual modules, and clearer attribution of errors, arguing that these benefits can outweigh those of end-to-end models. Many such points are well supported: for example, vehicle recognition can be handled efficiently with lightweight region filtering~\cite{masarykova2024single}, and geometric cues from camera frames can improve scene understanding~\cite{galinski2024ultra}. Nevertheless, existing modular pipelines have not yet demonstrated a practical way to counteract the propagation and accumulation of errors across stages when modules are integrated, while still meeting real-time computational constraints. Moreover, modular interfaces typically enforce compact information exchange, often passing only selected descriptors such as position, velocity, acceleration, distance, and related features.

\subsection{End-to-End solutions}

End-to-end formulations, by contrast, can carry informative cues about agent intent from perception to prediction, including signals such as turn indicators and brake lights, as well as body pose and other rich context. They are also appealing from an engineering standpoint due to single-loop training and simpler deployment. However, until recently, methods such as FaF~\cite{luo2018fastandfurious}, IntentNet~\cite{casas2018intentnet}, and PnPNet~\cite{liang2020pnpnet} did not fully realize the promise of end-to-end PnP, because they depended on intermediate feature-map representations and non-differentiable operations (e.g., non-maximum suppression). This design choice restricts error propagation between PnP components and often leads to only partial coupling through a composite loss.

Prior work~\cite{luo2018fastandfurious,casas2018intentnet,zeng2019neural_motion_planner} has nevertheless shown that agent intention can be inferred directly from raw sensor inputs. These approaches, however, performed direct prediction without incorporating explicit tracking, which weakened performance on occluded agents and limited the usable sensor history before model saturation.

Contemporary multimodal systems commonly build upon ideas introduced by PnPNet~\cite{liang2020pnpnet}. PnPNet established the feasibility of end-to-end trajectory prediction from raw sensor data for longer-horizon forecasting (\textgreater 1s), and it further emphasized the value of separating concerns while including explicit temporal tracking.

The introduction of Visual Trajectory Prediction via 3D agent queries (ViP3D)~\cite{gu2023vip3d} represented a notable step forward by demonstrating that end-to-end, fully differentiable prediction can be achieved in practice. Its query-based formulation provides a feature- and attention-driven mechanism that propagates perception information throughout the model. Additionally, by avoiding non-differentiable intermediate stages, it became the first fully differentiable vision-based trajectory prediction approach to use rich feature representations as the primary interface.

\begin{figure*}[!ht]
\centering
\includegraphics[width=\textwidth]{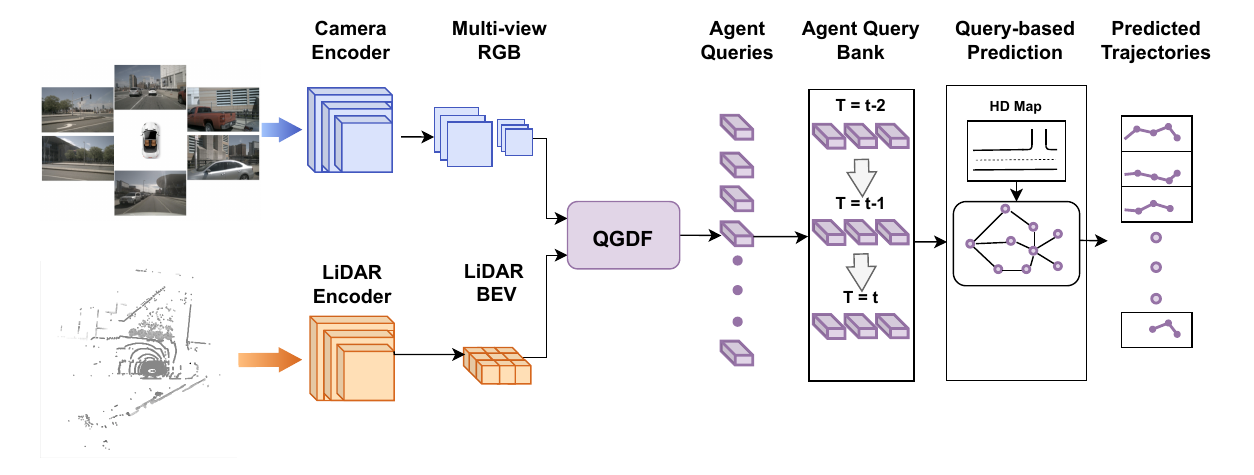}
\caption{Overall architecture of the Li-ViP3D++ multimodal end-to-end trajectory prediction model. The QGDF module is our primary contribution which handles sampling and fusion to ensure efficient usage of information from each input modality.} 
\label{fig:Li-ViP3D}
\end{figure*}
\subsection{Attention-based RGB and LiDAR fusion}
BEVFusion~\cite{liu2023bevfusion} presents an efficient strategy for lifting 2D camera observations into a 3D bird's-eye-view representation, enabling trajectory prediction to operate within a shared 3D space across modalities for most of the network. A key outcome is a reported 40$\times$ speedup over the prior standard BEV conversion technique, Lift-Splat-Shoot~\cite{philion2020lss} (LSS). The high cost of BEV conversion previously created a major bottleneck and hindered the development of BEV-centric methods, contributing to a gap in the literature.

Although BEVFusion moves toward more efficient multimodal fusion, it introduces a strong assumption: correct operation depends on sufficient spatial alignment across modalities. Recent methods such as LGMMFusion~\cite{cheng2025lgmmfusion} and RoboFormer~\cite{liu2024roboformer} show that integrating LiDAR can substantially improve agent-query-based models. At the same time, both approaches depend on explicit alignment in a shared BEV space. RoboFormer builds on the LSS lifting pipeline, whereas LGMMFusion introduces a specialized attention mechanism. Neither work explicitly discusses the additional latency introduced by the transformation.

Alternatively, simpler strategies~\cite{halinkovic2024intrinsically,brodermann2025cafuser} process the visual modalities independently, preserving spatial self-attention while enabling cross-modal interaction at the semantic level. This combination of spatial and encoder-level alignment can improve downstream representations efficiently and can also increase model explainability. In particular, encoder-level attention allows the contribution of each modality to be inspected, helping assess when a modality is informative and identify difficult edge cases for the network.

Our approach differs from these lines of work in two respects. First, unlike BEV-based fusion, it does not depend on constructing a shared dense BEV representation for cross-modal interaction at any point; instead, it uses sparse query-centered sampling from each modality at the predicted agent location. Second, unlike more generic deformable-attention fusion designs, the proposed QGDF explicitly separates modality-specific feature extraction from query-conditioned fusion, using a learned gate to determine the per-agent contribution of image and LiDAR evidence. This design makes the source of improvement more interpretable. The gain is primarily driven by better query-level modality selection and local feature acquisition rather than by heavier global fusion.

\section{Methodology}

We expand the Li-ViP3D architecture, a multimodal network that enables end-to-end perception and prediction based on agent queries and uses separate encoders for processing each input modality and an attention mechanism for query-level feature fusion. Our approach is designed to exploit multimodality and attention-based fusion to significantly increase the quality of detected and predicted agents at a minimal increase in computational cost. 

We use the large-scale nuScenes~\cite{caesar2020nuscenes} dataset to train and evaluate the model while following the standard train/validation split used in prior PnP work. The dataset contains 1000 urban scenes with a duration of 20 seconds each. In the default configuration, we utilize data from 6 RGB cameras sampled at 2 Hz, 32-beam LiDAR sweeps at 10 Hz, and an HD semantic map. Each input sample consists of 3 temporal frames of RGB and LiDAR data. Multi-view images are normalized using the ViP3D preprocessing convention and padded to a size divisor of 32, while LiDAR points are cropped to the predefined spatial range and voxelized before encoding. Training is performed for up to 24 epochs using AdamW with an initial learning rate of $10^{-4}$ and cosine annealing. In addition to this default setup, we also evaluate robustness under two controlled input modifications: removal of HD map input and reduction of RGB image resolution.

\subsection{Feature extraction}
The proposed multimodal model shown in Fig.~\ref{fig:Li-ViP3D} separately extracts 2D features from multi-view images, as proposed by DETR3D, using a ResNet50~\cite{resnet} as a camera encoder backbone, and processes multi-sweep LiDAR point cloud sweeps. We use 5 consecutive LiDAR sweeps (sampled at 10 Hz) spanning 0.5 s in a PointPillars~\cite{lang2019pointpillars} based LiDAR encoder. The ResNet50 encoder uses pretrained weights from DETR3D and we pretrained the PointPillars encoder for 3D detection on nuScenes. This configuration allows us to perform hybrid fusion with several benefits. In addition to improving 3D detection accuracy, we improve the temporal reasoning of the model by vastly increasing the temporal density of input data. This can be achieved using a computationally simple encoder. In our configuration, the PointPillars encoder operates on points within [-51.2,51.2] meters in  \textit{x/y} axes and [-5,3] meters in the \textit{z} axis. Points are discretized into pillars with voxel size (0.2,0.2,4.0), with up to 32 points per pillar, producing a dense BEV pseudo-image via pillar scattering. Each pillar is encoded by a single feed-forward network (FFN) layer (10-D point features → 64 channels) with BatchNorm and ReLU. The BEV map is then processed by a 3-stage 2D CNN backbone with stride 2 downsampling and channel widths {64, 128, 256}, followed by a Feature Pyramid Network (FPN) style upsampling neck (deconvolutions with strides 1/2/4) that merges multi-scale features into a 256-channel BEV representation used for subsequent multimodal fusion.

We utilize a tracked agent query memory bank as described by ViP3D~\cite{gu2023vip3d} and MOTR~\cite{zeng2022motr} to maintain information-dense historical feature states for each agent in the scene. Using agent queries, we can convert the feature representation to a space that allows for efficient object-level associations.

\subsection{Modality fusion}\label{sec:fusion}

\begin{figure}[!h]
\centering
\includegraphics[width=1\linewidth,height=17cm]{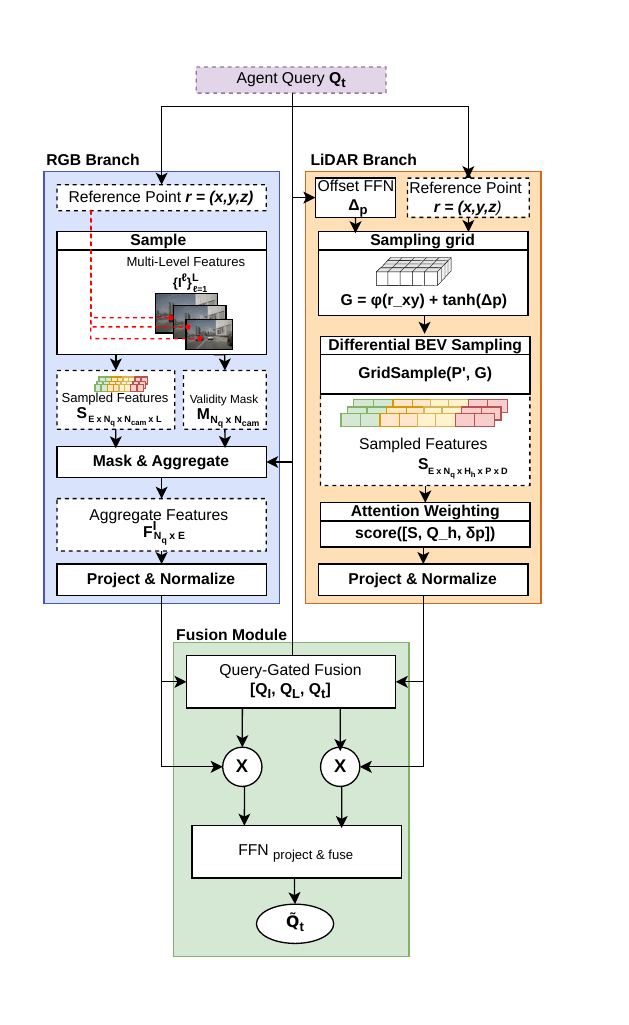}
\vspace{-2em}
\caption{Diagram of the Query-Gated Deformable Fusion (QGDF) mechanism and sampling from each modality.}
\label{fig:agent_feature_fusion}
\end{figure}

At each time step, we update the representation of every agent query using our attention-based QGDF module. The overall flow of the QGDF mechanism is shown in Fig.~\ref{fig:agent_feature_fusion}. QGDF lets each agent query gather evidence from the image and LiDAR branches at its estimated spatial location, and then adaptively decides how much each modality should contribute to the updated representation.

Let $\tens{Q}_t \in \mathbb{R}^{\nq \times B \times E}$ denote the set of agent queries at time step $t$, and let $r \in [0,1]^{B \times \nq \times 3}$ be their associated normalized 3D reference points. We denote the multi-level image features by $\{\tens{I}^{\ell}\}_{\ell=1}^{L}$, where each level $I^\ell \in \mathbb{R}^{B \times  \times E \times H_\ell \times W_\ell}$, and we denote the LiDAR BEV features (when available) by $P \in \mathbb{R}^{B \times E \times \hp \times \wpp}$.

\subsubsection{Image feature aggregation via masked attention over cameras and pyramid levels}
Rather than forming dot-product cross-attention between $\tens{Q}_t$ and dense image keys, we first sample image features at each query's reference point across all cameras and feature levels using a differentiable sampler. The sampling procedure projects $r$ into each camera using the provided calibration matrices and applies bilinear sampling in each pyramid level, producing sampled features: 
\begin{equation}
    \tens{S} = \Sample( \{I^\ell\}, \vect{r} ) \in \mathbb{R}^{B \times E \times \nq \times \ncam \times 1 \times L},
\end{equation}
The transformed reference points are also used to generate a validity mask:
\begin{equation}
    \tens{M} \in \{0,1\}^{B \times 1 \times \nq \times \ncam \times 1 \times 1},
\end{equation}
which indicates whether the projected reference point lies in front of the camera and within image bounds. We then predict per-query attention logits over the camera-level combinations from the query features:
\begin{equation}
    \tens{\omega}^{I} = \operatorname{FFN}_{I}(\tens{Q}_t) \in \mathbb{R}^{B \times \nq \times (\ncam L)}.
\end{equation}
To ensure that invalid views do not receive probability mass, we apply a masked softmax (using the expanded mask $M$ across pyramid levels) to obtain normalized weights $\tens{\alpha}^{I}$:
\begin{equation}
    \tens{\alpha}^{I} = \maskedsoftmax(\tens{\omega}^{I}, M) \in \mathbb{R}^{B \times 1 \times \nq \times \ncam \times 1 \times L}.
\end{equation}
Finally, we aggregate the sampled features across cameras and levels:
\begin{equation}
    F^{I} = \sum_{n=1}^{\ncam}\sum_{\ell=1}^{L} \tens{\alpha}^{I}_{n,\ell} \, S_{n,\ell}
    \in \mathbb{R}^{B \times E \times \nq}.
\end{equation}
This aggregation allows the image branch to ask which camera views and feature scales are most informative for this agent query, while ignoring views in which the projected point is not visible. The resulting image features are projected and normalized to align them with the query embedding space, yielding $ \tens{\bar{Q}}^{I}_t \in \mathbb{R}^{\nq \times B \times E}$.

\subsubsection{LiDAR feature aggregation via differentiable BEV sampling}

We extract LiDAR context for each query by differentiably sampling the BEV LiDAR feature map around the query's BEV reference location $r_{xy}$. Given a BEV feature map $P \in \mathbb{R}^{B \times \cl \times H \times W}$, we first project it into the transformer embedding space using a $1{\times}1$ convolution followed by spatial normalization, yielding features in $\mathbb{R}^{B \times E \times H \times W}$. This aligns the LiDAR representation with the transformer query embeddings and the image-derived features used in the fusion module.

For each query, we predict a small set of local 2D sampling offsets from its query embedding using a lightweight linear head:
\begin{equation}
    \Delta_P = \operatorname{FFN}(\tens{Q}_t)
    \in \mathbb{R}^{B \times \nq \times \hh \times \tens{p} \times 2},
\end{equation}
where $\hh$ denotes the number of sampling heads and $p$ denotes the number of sampled points per query and head. The BEV reference coordinate $r_{xy}$, normalized to $[0,1]$, is mapped to the $\texttt{grid\_sample}$ coordinate system in $[-1,1]$. The predicted offsets are bounded with a $\tanh$ nonlinearity and scaled to restrict sampling to a local neighborhood around the query reference point:
\begin{equation}
    \tens{G} = \operatorname{clip}\left(\phi(r_{xy}) + \tens{S} \cdot \tanh(\Delta_P), -1, 1\right),
\end{equation}
where $\phi(\cdot)$ denotes the transformation from normalized BEV coordinates to $\texttt{grid\_sample}$ coordinates, and $s$ controls the sampling radius.

We then sample the projected BEV feature map at the resulting per-query grid locations using bilinear interpolation with border padding:
\begin{equation}
    \tens{S} = \GridSample(P, G)
    \in \mathbb{R}^{B \times \nq \times \hh \times \tens{p} \times D},
\end{equation}
where $D$ is the per-head feature dimension and $E = \hh D$. This allows each query to collect nearby LiDAR evidence from a small learned BEV neighborhood instead of relying on a single rigid sampling location.

The sampled features are then aggregated using learned query-adaptive attention weights rather than being merely projected or averaged. For each sampled BEV feature, the attention score is conditioned on three sources of information: the sampled LiDAR feature itself, the corresponding per-head query embedding, and the relative sampling offset. This can be written as
\begin{equation}
    a_{q,h,p}
    =
    \operatorname{FFN}_{\mathrm{attn}}
    \left(
        \left[
            \tens{S}_{q,h,p};
            \tens{Q}_{q,h};
            \vect{\delta}_{q,h,p}
        \right]
    \right)
\end{equation}
where $S_{q,h,p}$ is the sampled LiDAR feature for query $q$, head $h$, and sampling point $p$, $Q_{q,h}$ is the corresponding query-head feature, $\delta_{q,h,p}$ is the relative sampling offset, and $[\space\cdot\space;\space\cdot\space]$ denotes concatenation. The resulting logits are normalized over the sampled points:
\begin{equation}
w_{q,h,p}
=
\operatorname{softmax}_{p}
\left(
    a_{q,h,p}
\right)
\end{equation}
The LiDAR context for each query and head is then computed as a weighted sum of the sampled features:
\begin{equation}
\tens{C}_{q,h}
=
\sum_{p=1}^{P}
w_{q,h,p}\tens{S}_{q,h,p}
\end{equation}

This attention-based aggregation allows the model to emphasize the most informative local BEV evidence for each query, while down-weighting less relevant samples from the learned neighborhood. The aggregated representation is then projected back to the transformer embedding space and normalized to match the image branch:
\begin{equation}
    \tens{\bar{Q}}^{L}_t
    = \BEVSample(P, r_{xy}, \tens{Q}_t)
    \in \mathbb{R}^{\nq \times B \times E}.
\end{equation}

Since the sampling is implemented with differentiable bilinear interpolation, gradients can flow to both the BEV feature map and the predicted sampling offsets. If LiDAR input is not provided, we set $\tens{\bar{Q}}^{L}_t$ to zero. This procedure replaces hard top-$k$ selection with continuous, query-adaptive BEV sampling and attention-based aggregation, yielding a stable LiDAR context vector per query that can be fused with the image-derived representation.

\subsubsection{Query-aware gated multimodal fusion}
The modality contributions are combined using a query-conditioned gating mechanism. We compute fusion logits from the concatenation of normalized modality features and the (detached) query representation:
\begin{equation}
    g_t = \operatorname{FFN}_{\text{gate}}\big([\tens{\bar{Q}}^{I}_t,\ \tens{\bar{Q}}^{L}_t, \tens{Q}_t]\big) \in \mathbb{R}^{\nq \times B \times 2},
\end{equation}
and obtain soft gates via
\begin{equation}
    \gamma_t = \operatorname{softmax}\!\left(g_t\right),
\end{equation}
The gated modality features are then formed as
\begin{equation}
    \hat{\tens{Q}}^{I}_t = \gamma_{t,0}\,\tens{\bar{Q}}^{I}_t,\qquad
    \hat{\tens{Q}}^{L}_t = \gamma_{t,1}\,\tens{\bar{Q}}^{L}_t,
\end{equation}
concatenated, and projected back to the query dimension:
\begin{equation}
    \tilde{\tens{Q}}_t = \operatorname{FFN}_{\text{proj\&fuse}}\big([\hat{\tens{Q}}^{I}_t,\ \hat{\tens{Q}}^{L}_t]\big) \in \mathbb{R}^{\nq \times B \times E}.
\end{equation}

The gate therefore acts as a query-specific confidence allocator, reducing the contribution of a modality in favor of the other modality if it is less informative for a particular agent.

\subsubsection{Residual update with positional encoding}
As in the original formulation, we incorporate a positional code derived from the reference points. Let $\phi(\cdot)$ denote a lightweight position encoder; we compute
\begin{equation}
    P_t = \phi\!\left(\invSigmoid(r)\right) \in \mathbb{R}^{\nq \times B \times E}.
\end{equation}
The updated query representation is then produced by a residual connection with dropout:
\begin{equation}
    Q_{t+1} = \tilde{\tens{Q}}_t + \tens{Q}_t + P_t.
\end{equation}

This procedure yields multimodally enhanced query representations that integrate camera evidence through masked attention over cameras and feature levels, LiDAR evidence through differentiable BEV sampling, and a query-conditioned gating mechanism that adaptively balances the two modalities.

\subsection{Agent Perception and Prediction}
Utilizing this multi-stage feature update strategy allows all downstream operations and subtasks to be performed on substantially richer query representations. At each time step $t$, every enhanced but currently unassigned query is passed through a query decoder that predicts the query centre coordinates $\hat{\vect{y}}_{\sigma(i)}$. These query-level predictions are then used to construct a bipartite matching problem, where the pairwise cost between a ground-truth target $y_i$ and a predicted query $\hat{\vect{y}}_{\sigma(i)}$ combines classification confidence and bounding-box distance:

\begin{align}
\mathcal{L}_{\text{match}}(y_i, \hat{y}_{\sigma(i)}) 
&= -\mathbf{1}_{\{c_i \ne \varnothing\}} \hat{p}_{\sigma(i)}(c_i) \nonumber \\
&\quad + \mathbf{1}_{\{c_i \ne \varnothing\}} \mathcal{L}_{\text{box}}(b_i, \hat{b}_{\sigma(i)})
\end{align}

Here, $\mathcal{L}_{\text{box}}$ denotes the $l_1$ loss over bounding-box parameters, $b_i$ is the ground-truth box, and $\hat{\vect{b}}_{\sigma(i)}$ together with $\hat{p}_{\sigma(i)}(c_i)$ correspond to the predicted box and the predicted probability of class $c_i$, respectively. Queries that have already been matched to a ground-truth agent preserve this assignment across time, unless the query is predicted as empty. Such persistent assignments are removed from the pairwise cost matrix and are therefore not reconsidered during matching.

To carry temporal context through the query memory bank, we additionally apply a temporal cross-attention operation. For each agent query $\tens{Q}_t^i$, we attend to its own stored history in the memory bank and compute a compact temporal summary:
\begin{equation}
\tilde{\vect{q}}_{t}^{\,i}
=
\operatorname{softmax}
\left(
\frac{
\vect{q}_{t,\mathrm{query}}^{\,i}
\left(\tens{Q}_{\mathrm{bank,key}}^{\,i}\right)^{\top}
}
{\sqrt{d}}
\right)
\tens{Q}_{\mathrm{bank,value}}^{\,i}.
\end{equation}

Following MOTR~\cite{zeng2022motr}, the query memory bank is implemented as a first-in-first-out queue. We store a fixed history of four previous time steps per query. This design enables efficient agent-level temporal processing by compressing the historical states associated with query $q^i$ contained in $Q^i_{\text{bank}}$. Importantly, the temporal aggregation is performed independently for each query to improve computational efficiency and to prevent information leakage across different agents. The resulting query feature is obtained using a two-layer feedforward network:
\begin{equation}
\vect{q}_{t}^{\,i'}
=
\operatorname{FFN}
\left(
\vect{q}_{t}^{\,i} + \tilde{\vect{q}}_{t}^{\,i}
\right),
\end{equation}
thereby integrating the spatio-temporal information produced by the multimodal feature update stages.

For trajectory forecasting, we follow the established design choices of ViP3D in the default configuration. The enhanced agent queries are further processed together with an HD semantic map encoded using VectorNet~\cite{gao2020vectornet}, which provides a vectorized representation of contextual road semantics, including lane structure, traffic signs, and related map elements. Final trajectory outputs are generated using a regression-based decoder implemented as a two-layer FFN, predicting $K$ candidate trajectories per agent. For the map-removal robustness experiments described in Section~\ref{sec:robustness_protocol}, this map-based interaction stage is replaced by a map-free interaction module while keeping the remaining prediction pipeline unchanged.

\begin{table*}[!ht]
\centering
\caption{Performance comparison between Traditional, PnPNet-vision, ViP3D and our proposed method. Lower is better for minADE, minFDE, and MR; higher is better for EPA.}
\begin{tabular}{lccccc}
\toprule
\textbf{Metrics} & \textbf{Det3D + Kalman Filter} & \textbf{PnPNet-vision~\cite{liang2020pnpnet}} & \textbf{ViP3D~\cite{gu2023vip3d}} & \textbf{Li-ViP3D~\cite{halinkovic2025LiViP3D}} & \textbf{Li-ViP3D++ (Ours)} \\
\midrule
$\mathrm{minADE}_{k}\downarrow$ & 2.07 m & 2.04 m & 1.52 m & 1.45 m & \textbf{1.11 m}\\
$\mathrm{minFDE}_{k}\downarrow$ & 3.10 m & 3.08 m & 2.35 m & 2.20 m & \textbf{1.80 m}\\
$\mathrm{MR}$$\downarrow$     & 0.289 & 0.277 & 0.239 & 0.236 & \textbf{0.209}\\
$\mathrm{EPA}$$\uparrow$      & 0.191 & 0.198 & 0.236 & 0.250 & \textbf{0.505}\\
$\mathrm{FP~ratio}$$\downarrow$      & - & - & 0.231 & 0.221 & \textbf{0.069}\\
$Recall$$\uparrow$      & - & - & 0.469 & 0.458 & \textbf{0.571}\\
$Precision$$\uparrow$      & - & - & 0.769 & 0.779 & \textbf{0.931}\\
$\mathrm{mAP}$$\uparrow$      & - & - & 0.472 & 0.472 & \textbf{0.616}\\
\bottomrule
\end{tabular}
\label{tab:performance_comparison}
\end{table*}

Consistent with the end-to-end, fully differentiable formulation, the network is trained with a joint objective that optimizes classification, bounding-box regression, and trajectory decoding simultaneously. The classification and bounding-box regression terms are computed as:

\begin{equation}
\mathcal{L}_{\text{cls}} = \sum_{i=1}^{N} -\log \hat{p}_{\hat{\sigma}(i)}(c_i),
\end{equation}

\begin{equation}
\mathcal{L}_{\text{coord}} = \sum_{i=1}^{N} \mathbf{1}_{\{c_i \ne \varnothing\}} \mathcal{L}_{\text{box}}(b_i, \hat{b}_{\hat{\sigma}(i)}).
\end{equation}

The regression-based trajectory decoding loss is defined as an $l_1$ regression term, $\mathcal{L}_{\text{reg}}$, computed between the ground-truth future trajectory $s$ and the best-matching predicted hypothesis $s^{(\hat{k})}$. Concretely, we evaluate the loss over the prediction horizon $\tfuture$ as

\begin{equation}
\mathcal{L}_{\text{trajectory}} = \sum_{t=1}^{\tfuture} \mathcal{L}_{\text{reg}}(s_t, s_t^{(\hat{k})}).
\end{equation}

The index $\hat{k}$ is selected using a minimum-distance criterion: among the $K$ predicted trajectories, we choose the one that minimizes the $l_2$ distance to the ground truth when comparing all corresponding trajectory points. With this selection rule, the overall training objective becomes

\begin{equation}
\mathcal{L} = \mathcal{L}_{\text{cls}} + \mathcal{L}_{\text{coord}} + \mathcal{L}_{\text{trajectory}}.
\end{equation}

\subsection{Robustness evaluation protocol}\label{sec:robustness_protocol} 

In addition to evaluating the default architecture, we assess robustness under reduced infrastructure and degraded visual input conditions. We consider three modified settings: (i) \textit{without HD map}, where the map-based forecasting context is removed; (ii) \textit{reduced resolution}, where the input RGB images are downsampled from the default 1600$\times$900 pixels to 900$\times$544 pixels; and (iii) \textit{without HD map + reduced resolution}, where both modifications are applied simultaneously.

\begin{figure}[!h]
    \centering
    \includegraphics[width=1\linewidth]{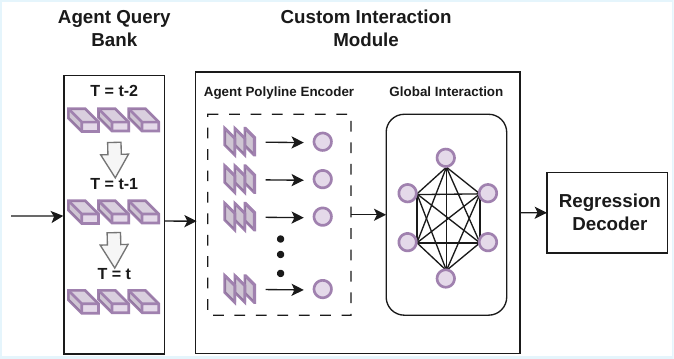}
    \caption{Custom graph-based interaction module. Agent query embeddings across all time steps are encoded into a unified representation which is used for calculating global interaction.}
    \label{fig:custom_interaction_module}
\end{figure}

For experiments without HD map input, we replace the VectorNet-based map interaction stage with a lightweight map-free interaction module that operates directly on the enhanced agent queries and their temporal histories, as depicted in Figure~\ref{fig:custom_interaction_module}, replacing the Query-based
Prediction module as seen in Figure~\ref{fig:Li-ViP3D}. The interaction module replicates the VectorNet formula while removing any non-query elements. First, it encodes all query embeddings using a self-attention mechanism into a single representation per agent. Another self-attention mechanism is then used to model the global interaction between each agent representation. This module models agent-agent interactions without relying on external semantic map priors, allowing us to evaluate whether the proposed fusion mechanism remains effective when structured map context is unavailable. The trajectory decoder and all other parts of the perception-and-prediction pipeline remain unchanged.

For the reduced-resolution experiments, only the spatial resolution of the RGB input is modified; the model architecture, training setup, LiDAR configuration, and evaluation protocol are kept the same. This allows us to isolate the effect of weaker visual input quality on multimodal query fusion. All robustness experiments are evaluated using the same metrics as the default configuration.

\begin{table*}[!ht]
\centering
\caption{Performance comparison of ViP3D, Li-ViP3D, and Li-ViP3D++ under various conditions. As "Original Inputs" the models use 1600 $\times$ 900 px RGB images and HD map encoding provided by VectorNet. The Reduced Resolution experiment downscales input RGB images to 900 $\times$ 544 px. Experiments without HD map use our custom graph interaction/prediction module instead of VectorNet. Best results for the given metric are highlighted in \textbf{bold}, the second best results are in \underline{underlined}.}
\small
\begin{tabularx}{\textwidth}{l *{8}{>{\centering\arraybackslash}X}}
\toprule
\textbf{Models\textbackslash Metrics}
& \textbf{minADE $\downarrow$}
& \textbf{minFDE $\downarrow$}
& \textbf{MR $\downarrow$}
& \textbf{EPA $\uparrow$}
& \textbf{FP ratio $\downarrow$}
& \textbf{Recall $\uparrow$}
& \textbf{Precision$\uparrow$}
& \textbf{mAP $\uparrow$}  \\
\midrule
\multicolumn{9}{l}{\textbf{Original Inputs}} \\
\midrule
ViP3D      & 1.52 m & 2.35 m & 0.239 & 0.236 & 0.231 & 0.469 & 0.769 & 0.472 \\
Li-ViP3D   & 1.45 m & 2.20 m & 0.236 & 0.250  & 0.221 & 0.458 & 0.779 & 0.472 \\
Li-ViP3D++ (Ours)& \textbf{1.11 m} & \textbf{1.80 m} & \underline{0.209} & \textbf{0.505} & \textbf{0.069} & \underline{0.571} & \textbf{0.931} & 0.616 \\
\addlinespace[3pt]
\midrule
\multicolumn{9}{l}{\textbf{Without HD map}} \\
\midrule
ViP3D      & 1.57 m & 2.38 m & 0.241 & 0.239 & 0.232 & 0.462 & 0.760  & 0.474 \\
Li-ViP3D    & 1.54 m & 2.42 m & 0.242 & 0.226 & 0.242 & 0.468 & 0.758 & 0.473 \\
Li-ViP3D++ (Ours)  & \underline{1.13 m} & 1.87 m & \underline{0.209} & \underline{0.488} & 0.085 & 0.570 & 0.915 & \underline{0.622} \\
\addlinespace[3pt]
\midrule
\multicolumn{9}{l}{\textbf{Reduced Resolution}} \\
\midrule
ViP3D      & 1.54 m & 2.45 m & 0.275 & 0.148 & 0.270  & 0.403 & 0.73  & 0.446 \\
Li-ViP3D    & 1.51 m & 2.32 m & 0.257 & 0.180  & 0.259 & 0.426 & 0.741 & 0.447 \\
Li-ViP3D++ (Ours)  & 1.14 m & \underline{1.84 m} & \textbf{0.204} & 0.477 & 0.107  & \textbf{0.587} & 0.892 & \textbf{0.631} \\
\addlinespace[3pt]

\midrule
\multicolumn{9}{l}{\textbf{Without HD map and Reduced Resolution}} \\
\midrule
ViP3D      & 1.51 m & 2.32 m & 0.255 & 0.193 & 0.247 & 0.420  & 0.753 & 0.460  \\
Li-ViP3D    & 1.50 m & 2.38 m & 0.250  & 0.192 & 0.255 & 0.437 & 0.745 & 0.457 \\
Li-ViP3D++ (Ours)  & \underline{1.13 m} & 1.93 m & 0.219 & 0.481 & \underline{0.081} & 0.559 & \underline{0.918} & 0.615 \\
\bottomrule
\end{tabularx}

\label{tab:experimental_comparisons}
\end{table*}

\section{Results}
We evaluate the proposed fusion mechanism using a set of complementary detection and prediction metrics, selected to provide a robust view of overall system performance and to isolate the contribution of the fusion strategy.

For trajectory forecasting, we report minimum Average Displacement Error ($minADE_k$) and minimum Final Displacement Error ($minFDE_k$), which quantify the mean and final $l_2$ distance, respectively, between predicted and ground-truth agent positions under the best-matching hypothesis among $K$ outputs. We also include the Miss Rate ($\mathrm{MR}$), defined here as the fraction of predictions whose $minFDE$ exceeds 2 meters. To capture end-to-end behavior in the presence of false positives, we also report End-to-end Prediction Accuracy (EPA)~\cite{gu2023vip3d}. EPA explicitly accounts for false positives by matching predicted agents to ground truth based on spatial proximity and computing a normalized hit rate as the ratio of matched predictions to the total number of ground-truth agents.

As displacement errors and derived metrics are notoriously insufficient for evaluating end-to-end PnP models due to the low impact of incorrect detections on final results, we include a set of detection-focused metrics. These include $Precision$, $Recall$, $\mathrm{mAP}$, and $FP\ ratio$ (normalized average ratio of false positive detections per scene). It is important to note that due to the end-to-end nature of the evaluated architectures, the detections are evaluated at the first predicted timestep.

All experiments are conducted within a region spanning 51.2 meters in each direction and cover seven agent categories: car, truck, bus, trailer, motorcycle, bicycle, and pedestrian. Trajectories are evaluated over a forecasting horizon of $\Delta T = 6$ seconds, and the model outputs $K = 6$ candidate future trajectories per agent.

The primary comparison between our method and the considered baselines is reported in Table~\ref{tab:performance_comparison}. We evaluate against (i) a conventional modular pipeline that separates Perception and Prediction, (ii) a modified vision-based PnPNet variant, (iii) ViP3D, and (iv) the original Li-ViP3D. For consistency and to ensure a fair comparison, all methods employ a regression-based trajectory decoder. We also retrained the ViP3D architecture with the same setup and hardware as the multimodal architectures to make sure the comparisons are as accurate as possible. The retrained version achieved better displacement error metrics, $\mathrm{minADE}_{k}$ of 1.52 meters and $\mathrm{minFDE}_{k}$ of 2.35 meters, compared to the results published in the ViP3D paper ($\mathrm{minADE}_{k}$ of 2.03 meters and $\mathrm{minFDE}_{k}$ of 2.90 meters).

The proposed improved fusion mechanism for enhancement of agent queries leads to significant improvements in trajectory prediction metrics, specifically displacement error and miss rate performance, compared to the Li-ViP3D and retrained ViP3D architectures for the $\Delta T = 6$ seconds prediction horizon, as used in the ViP3D evaluation protocol. Our Li-ViP3D++ architecture also achieves significantly better detection results on key metrics and end-to-end prediction accuracy.

The proposed QGDF fusion achieves an absolute 25.5\% increase in EPA, the most robust prediction metric. This large improvement stems from a significant reduction in false positive detections. The ratio of hallucinated agents drops from an average of 22.1\% in Li-ViP3D to 6.9\%. This improvement tracks across all other detection metrics ($Precision$, $Recall$, and $\mathrm{mAP}$), signifying the impact of our architectural improvements on the quality of detection outputs. 

\subsection{Robustness under reduced infrastructure and visual quality} 
We further evaluate the proposed fusion mechanism under reduced infrastructure and degraded visual input. Table~\ref{tab:experimental_comparisons} compares the performance of all models in the original configuration and in three modified settings: without HD map input, with reduced RGB resolution, and with both constraints applied simultaneously.

Across all settings, Li-ViP3D++ remains the strongest model in all metrics. While predictably, the Li-ViP3D++ variant with access to full resolution input and HD map is dominant across most metrics, the other variants are not far behind. Under reduced-resolution RGB input, it reaches the best results for miss rate (0.204), Recall (0.587), and mAP (0.631). Even when both constraints are combined, Li-ViP3D++ preserves strong performance, achieving second-best results for minADE (1.13 m), False Positive ratio (0.081) and Precision (0.918). This variant still outperforms the ViP3D and Li-ViP3D models without constraints on all metrics. 

These results indicate that the proposed fusion mechanism remains effective even when external map support is removed and visual input quality is reduced. In particular, the main benefit of improved query grounding and reduced hallucinated detections persists across all evaluated conditions.

\subsection{Per-class performance}

As all of the query-based architectures discussed in this paper focus on multi-agent prediction, we also provide Table~\ref{tab:per_class_epa_map} to better understand how the improved fusion mechanism affects individual agent categories. The table reports per-class EPA and mAP for the original input settings. We focus on EPA and mAP, as they represent the most important prediction and classification metrics, respectively.

\begin{table}[!h]
\centering
\small
\setlength{\tabcolsep}{3pt}
\caption{Per-class EPA and mAP comparison under original input settings. Best results are shown in \textbf{bold}.}
\label{tab:per_class_epa_map}
\begin{tabular}{lcccccc}
\toprule
\multirow{2}{*}{\textbf{Class}} &
\multicolumn{2}{c}{\textbf{ViP3D}} &
\multicolumn{2}{c}{\textbf{Li-ViP3D}} &
\multicolumn{2}{c}{\textbf{Li-ViP3D++}} \\
\cmidrule(lr){2-3}\cmidrule(lr){4-5}\cmidrule(lr){6-7}
& \textbf{EPA} & \textbf{mAP} & \textbf{EPA} & \textbf{mAP} & \textbf{EPA} & \textbf{mAP} \\
\midrule
Car        & 0.277 & 0.590 & 0.295 & 0.590 & \textbf{0.532} & \textbf{0.717} \\
Truck      & 0.104 & 0.324 & 0.085 & 0.354 & \textbf{0.398} & \textbf{0.515} \\
Bus        & 0.079 & 0.342 & 0.080 & 0.335 & \textbf{0.384} & \textbf{0.619} \\
Trailer    & 0.000 & 0.104 & 0.000 & 0.114 & \textbf{0.131} & \textbf{0.227} \\
Motorcycle & 0.135 & 0.268 & 0.128 & 0.278 & \textbf{0.273} & \textbf{0.355} \\
Bicycle    & 0.083 & 0.296 & 0.054 & 0.315 & \textbf{0.187} & \textbf{0.325} \\
Pedestrian & 0.220 & 0.311 & 0.194 & 0.305 & \textbf{0.413} & \textbf{0.433} \\
\bottomrule
\end{tabular}
\end{table}

Li-ViP3D++ achieves the best EPA and mAP for every evaluated class, indicating that the proposed query-guided fusion improves both trajectory prediction quality and detection reliability across the full class distribution. The largest EPA gains are observed for large dynamic objects such as trucks and buses, where the multimodal fusion provides substantially better agent grounding. The method also improves performance for more challenging and less frequent categories, including trailers, motorcycles, and bicycles, although their absolute scores remain lower than those of cars and pedestrians. The differences in inter-class performance can largely be explained by the class imbalances in the nuScenes dataset. For example, the weakest results are achieved for the \textit{Trailer} and \textit{Bicycle} classes which are represented by only 2\% and 1\% of the provided annotations respectively, while the \textit{Car} class represents 42\% of the provided annotations.

\subsection{Ablation study}

To isolate the contribution of the individual components of the proposed Li-ViP3D++ fusion mechanism, we conduct an ablation study under the original input setting. Table~\ref{tab:ablation_study} reports the performance of the full Li-ViP3D++ model together with three ablated variants: a gating ablation (no learnable modality weighting), an offset refinement ablation (fixed sampling point), and a full ablation where both components are removed.

\begin{table}[!h]
\centering
\small
\setlength{\tabcolsep}{4pt}
\caption{Ablation study of the proposed Li-ViP3D++ fusion mechanism under original input settings. Best results for each metric are shown in \textbf{bold}.}
\label{tab:ablation_study}
\begin{tabular}{lcccc}
\toprule
\textbf{Metric} & \textbf{Li-ViP3D++} & \textbf{- gating} & \textbf{- offset} & \textbf{- gating \& offset} \\
\midrule
minADE $\downarrow$      & \textbf{1.11} & 1.13 & 1.17 & 1.20 \\
minFDE $\downarrow$      & \textbf{1.80} & 1.87 & 1.93 & 2.03 \\
MR $\downarrow$          & \textbf{0.209} & 0.215 & 0.232 & 0.231 \\
EPA $\uparrow$           & \textbf{0.505} & 0.487 & 0.464 & 0.471 \\
FP ratio $\downarrow$    & \textbf{0.069} & 0.077 & 0.080 & 0.086 \\
Recall $\uparrow$        & 0.571 & \textbf{0.582} & 0.538 & 0.555 \\
Precision $\uparrow$     & \textbf{0.931} & 0.923 & 0.920 & 0.914 \\
mAP $\uparrow$           & 0.616 & \textbf{0.624} & 0.589 & 0.622 \\
\bottomrule
\end{tabular}
\end{table}

The ablation results show that the complete Li-ViP3D++ model achieves the best overall performance when it comes to end-to-end behavior. Removing the offset refinement and forcing a single fixed sampling position leads to a clear degradation in both trajectory quality and end-to-end prediction performance, increasing minADE from 1.11 to 1.17 and minFDE from 1.80 to 1.93, while also reducing EPA and mAP. Interestingly, it still presents a strong improvement over the original Li-ViP3D model, demonstrating that a single sampling point provides more usable information than an aggregate of top-$k$ nearest points which the model is unable to explicitly filter. The full ablation further degrades displacement accuracy to 1.20 minADE and 2.03 minFDE, confirming that both components contribute to improved forecasting quality.

The gating ablation, which forces equivalent importance for both modalities, achieves slightly higher Recall and mAP than the full model. However, this should not be interpreted as a failure of the gating mechanism. The full Li-ViP3D++ model is not optimized to maximize every detection metric independently; rather, it produces the best overall end-to-end behavior by prioritizing more selective query grounding, higher Precision, fewer false positives, stronger EPA, and better trajectory forecasting. In this sense, the learned gate shifts the model toward more reliable and better-calibrated detections, even if a non-gated variant can recover a slightly larger number of candidate objects. This suggests that the gating mechanism primarily improves the quality and selectivity of the fused agent queries, leading to more accurate and better-grounded trajectory estimates. The gating mechanism presents a clear shift towards increased selectivity, indicating that dynamic modality reliance is important especially in edge cases where the modalities provide conflicting information. The impact of the learnable modality gate is further discussed in Section~\ref{sec:disc_modality}

\subsection{Efficiency evaluation}\label{sec:efficiency}
\begin{table}[!h]
\centering
\caption{Latency comparison (mean \(\pm\) std) across methods, measured in milliseconds across 1000 independent inferences per model. Lower is better.}
\label{tab:latency}
\begin{tabular}{lcccc}
\toprule
Method & Mean & Std & Min & Max \\
\midrule
\multicolumn{5}{l}{\textbf{Original Inputs}} \\
\midrule
ViP3D & 117.27 & 0.06 & 117.19 & 117.36 \\
Li-ViP3D++ & 139.82 & 0.14 & 139.64 & 140.10 \\
Li-ViP3D & 145.91 & 0.36 & 145.56 & 146.69 \\
\midrule
\multicolumn{5}{l}{\textbf{Without HD map and Reduced Resolution}} \\
\midrule
ViP3D & 52.51 & 0.05 & 52.44 & 52.62 \\
Li-ViP3D++ & 75.40 & 0.31 & 75.14 & 76.28 \\
Li-ViP3D & 79.85 & 0.29 & 79.48 & 80.29 \\
\bottomrule
\end{tabular}
\end{table}

\begin{figure*}[!ht]
\centering
\includegraphics[width=\textwidth]{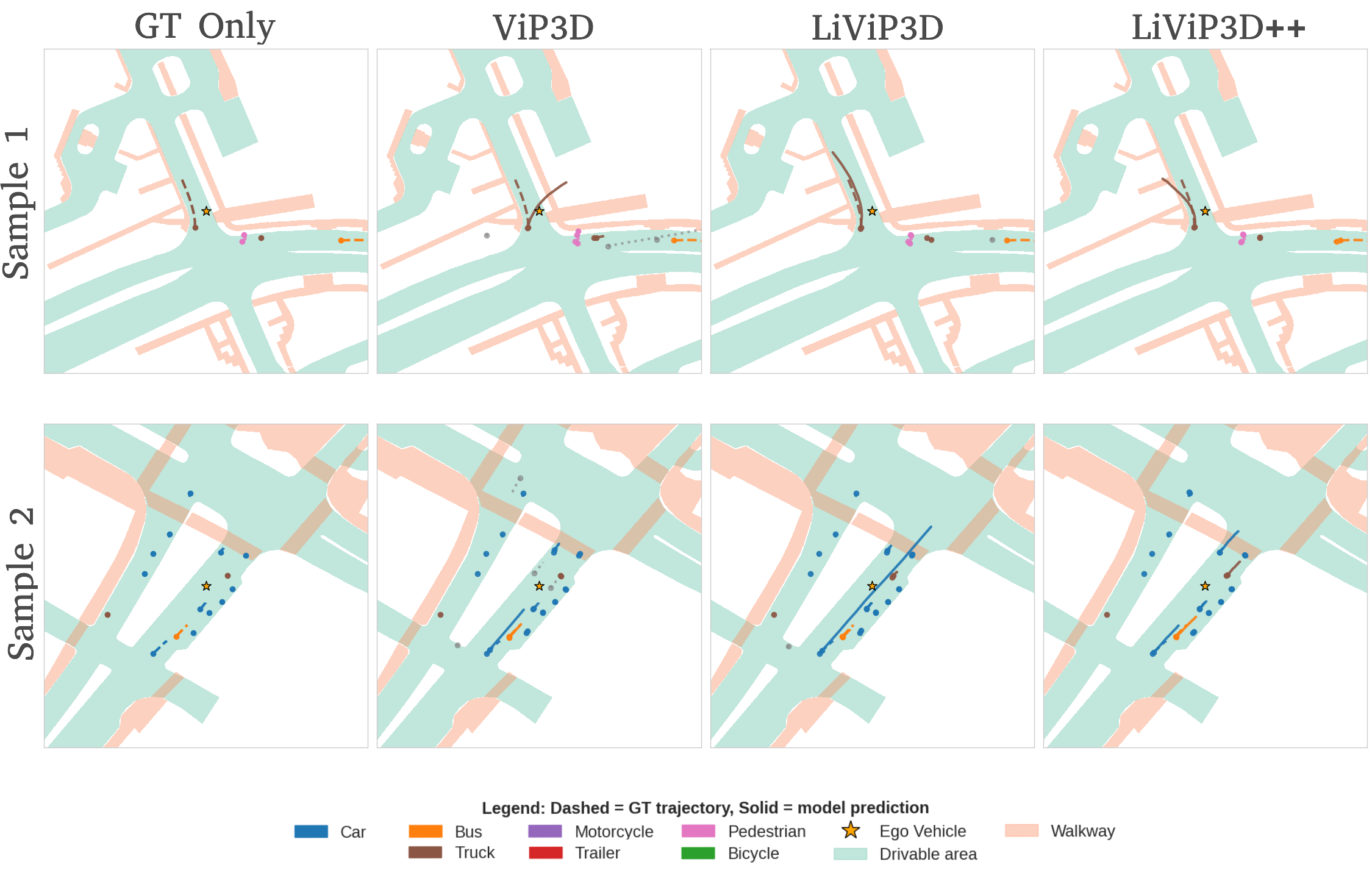}
\caption{Visual comparisons of the predicted trajectories for each of the compared agent query-based models. Each model predicted $K=6$ possible future trajectories, we visualize the trajectory with the highest confidence for each sample.}
\label{fig:prediction_visualization}
\end{figure*}

Beyond quantitative performance, it is important for PnP to consider the time constraints of their applications. Table~\ref{tab:latency} reports the latency comparison of our end-to-end Li-ViP3D++ architecture. All compared architectures used ResNet50 as their image backbone, the multimodal architectures both used PointPillars with the same configuration as the LiDAR backbone. In the default setting, the RGB input resolution is 1600 $\times$ 900 pixels.

Under the default input configuration, the proposed Li-ViP3D++ architecture is slightly faster than the original Li-ViP3D implementation (139.82 ms vs.\ 145.91 ms), corresponding to a reduction of 6.09 ms or approximately 4.4\%. Although Li-ViP3D++ remains slower than ViP3D (117.27 ms), this gap is expected given the added multimodal processing and query-level fusion. Importantly, the improved end-to-end quality of Li-ViP3D++ is therefore not obtained at the cost of increased runtime relative to the original multimodal baseline.

The constrained setting without HD map input and with reduced RGB resolution leads to a much larger latency reduction for all evaluated models. In this setting, Li-ViP3D++ decreases from 139.82 ms to 75.40 ms, which corresponds to a 46.1\% reduction in inference time. ViP3D decreases from 117.27 ms to 52.51 ms, corresponding to a 55.2\% reduction. These results show that a large fraction of the computational cost can be removed while preserving strong end-to-end prediction performance, especially for the proposed multimodal architecture. The majority of latency reduction in these experiments came from the reduced resolution of RGB inputs. In isolation, the replacement of VectorNet with our custom interaction module only led to an insignificant latency reduction of 1--5 ms on average across the architectures.

All latency measurements were obtained on a single NVIDIA GeForce RTX 4090 GPU and AMD Ryzen Threadripper PRO 7965WX CPU. No inference-time optimizations, such as TensorRT, were applied to any of the evaluated models. 

\section{Discussion}

The results highlight a clear shift in the performance profile induced by the proposed fusion mechanism. Compared to the original Li-ViP3D, the proposed Li-ViP3D++ improves both trajectory accuracy and end-to-end behavior. Across the evaluated settings, Li-ViP3D++ achieves superior displacement-error metrics, miss rate, EPA, and detection quality relative to both Li-ViP3D and ViP3D. This improvement is particularly evident when comparing standard forecasting metrics (minADE/minFDE/MR) against EPA and the explicit detection metrics. In the original input setting, Li-ViP3D++ moderately reduces minADE from 1.45 m to 1.11 m and minFDE from 2.20 m to 1.80 m compared to Li-ViP3D. However, thanks to the overall improvement in end-to-end behavior, EPA more than doubles from 0.250 to 0.505. At the same time, the false positive ratio is substantially reduced, indicating that the model is not only more accurate in its trajectory forecasts, but also more reliable in determining which agents exist and should be tracked. This behavior is consistent with the definition of EPA, which jointly evaluates detection and forecasting rather than measuring trajectory error only for already matched predictions. As detection and query grounding improve, the model is able to recover a broader and more accurate set of agents without degrading trajectory quality. In practical autonomy stacks, this behavior is highly desirable. Hallucinated agents can trigger unnecessary braking or planning failures even if the predicted trajectories for true agents are accurate. 

A key observation is that the benefits are consistent across categories and are reflected in both precision-oriented metrics (Precision, FP ratio) and recall-oriented metrics (Recall, mAP). Li-ViP3D++ improves over the baselines across all reported evaluation conditions, including the original inputs, removal of the HD map, reduced RGB resolution, and the combined constrained setting. Although recall remains moderate across all experiments, with a best value of 0.587 across all experiments (Li-ViP3D++ with reduced RGB resolution), this should be interpreted in light of the end-to-end evaluation protocol, where detections are assessed at the first predicted timestep. Under this setting, difficult or late-emerging agents are more likely to be counted as missed detections, so the recall values reflect the strictness of the protocol rather than an isolated weakness of the proposed architecture. Importantly, the proposed model still achieves the strongest recall among the compared architectures in the evaluated settings, while also maintaining substantially higher precision and lower false positive ratios. This suggests that the new query enhancement does not merely suppress detections indiscriminately; rather, it improves the semantic and spatial alignment between predicted queries and real scene entities.

The per-class results further confirm the trend of across-the-board improvements. Li-ViP3D++ achieves the best EPA and mAP for every evaluated class, showing that the improvements are not concentrated in a single dominant category but extend across cars, trucks, buses, trailers, motorcycles, bicycles, and pedestrians. While the performance of Li-ViP3D++ for the \textit{Trailer} and \textit{Bicycle} classes remains modest despite the improvements, it can largely be attributed to insufficient representations in training data. The practical impacts of the proposed approach are highlighted by the efficiency experiments described in Section~\ref{sec:efficiency}. Despite reduced input resolution and removed HD map support, Li-ViP3D++ outperforms both baselines without these constraints on all metrics while reducing inference latency by 36\% compared to the unimodal ViP3D and 48\% compared to the multimodal Li-ViP3D.

\subsection{Qualitative analysis}

Qualitatively, we observe that the enhanced fusion mechanism produces a clear trend toward better query grounding. The model outputs, together with the ground truth and baseline predictions, are shown in Fig.~\ref{fig:prediction_visualization} for two representative scenes. As the quantitative metrics suggest, Li-ViP3D++ produces fewer complete hallucinations and fewer predictions that are marked as false positives because their initial positions are more than 2 m away from the corresponding ground-truth agent.

To better characterize this effect, we further analyzed the false positives produced by ViP3D, the original Li-ViP3D, and the proposed Li-ViP3D++. We grouped false positives into three diagnostic categories: duplicate tracks, mislocalized predictions near ground-truth agents (between 2 and 4 meters), and ghost agents located more than 4 meters from any ground-truth agent. Compared with ViP3D, Li-ViP3D++ reduces the total number of false positives from 18,034 to 5,236. The largest contribution to this reduction comes from mislocalized predictions near ground-truth agents, which decrease from 9,132 to 1,349. This accounts for slightly more than half of the total false-positive reduction, indicating that QGDF primarily improves the spatial grounding of agent queries rather than simply suppressing detections. A similar trend is visible relative to the original Li-ViP3D, where mislocalized predictions decrease from 8,676 to 1,349.

The reduction is not limited to mislocalized predictions. Ghost agents are also reduced substantially, from 6,264 in ViP3D and 5,754 in the original Li-ViP3D to 2,921 in Li-ViP3D++, corresponding to an approximately two-fold reduction. Duplicate tracks are reduced from 2,638 in ViP3D and 2,230 in the original Li-ViP3D to 966 in Li-ViP3D++, leaving roughly one third of the ViP3D count and less than half of the original Li-ViP3D count. These trends suggest that QGDF removes false positives across multiple failure modes. It mainly reduces weakly grounded or spatially misaligned queries, while also suppressing fully hallucinated agents and redundant duplicate tracks.

Overall, the proposed model's predictions are reasonable and internally consistent. The model can handle more complex maneuvers while remaining within expected bounds based on the scene -- vehicle agents tend to remain within drivable areas while pedestrians remain in walkways and crossings. We also observe an increase in intra-agent consistency within scenes. As can be seen in Fig.~\ref{fig:prediction_visualization}, Sample 2, the model expects neighboring agents of the same class to be moving at a similar velocity. This leads to more consistent trajectories and fewer outliers. Overall, the new fusion mechanism leads to a better utilization of LiDAR data and thus a decrease in seemingly random over-predictions of trajectory length.

\subsection{Modality usage and reliance}\label{sec:disc_modality}
The learnable modality fusion mechanism described in Section~\ref{sec:fusion} lets us analyze the inferred reliance on LiDAR features within the enhanced queries. Overall, we observe that the model learned to utilize the LiDAR modality successfully. The mean usage of the LiDAR modality across all predicted agents in our testing data was 45.7\%, almost equaling the importance of RGB data despite the lightweight nature of the LiDAR branch.

While the usage ratio of both modalities remains relatively stable across all scenes, two trends are notable. First, reliance on LiDAR varies by class: for example, the \textit{truck} and \textit{trailer} class exhibit an average LiDAR contribution of 43\%, while motorcycles/bicycles show a higher contribution (47\%). One plausible explanation is that smaller agents are more sensitive to appearance ambiguity and partial occlusions in RGB, so even modest LiDAR evidence can provide a stabilizing geometric cue that helps disambiguate true objects from clutter while the larger \textit{truck} and \textit{trailer} are rarely ever significantly occluded in the RGB source data.

Second, the LiDAR contribution increases with point support. While outlier queries with fewer than 5 LiDAR points in the predicted bounding box area still show LiDAR reliance at an average of 44\%, queries supported by more than 100 points rise to 48.5\%. Every other measured bin (5-10, 10-20, ..., 90-100) showed a reliance between 45-46\%. 

\begin{figure}[!h]
\centering
\includegraphics[width=1\linewidth]{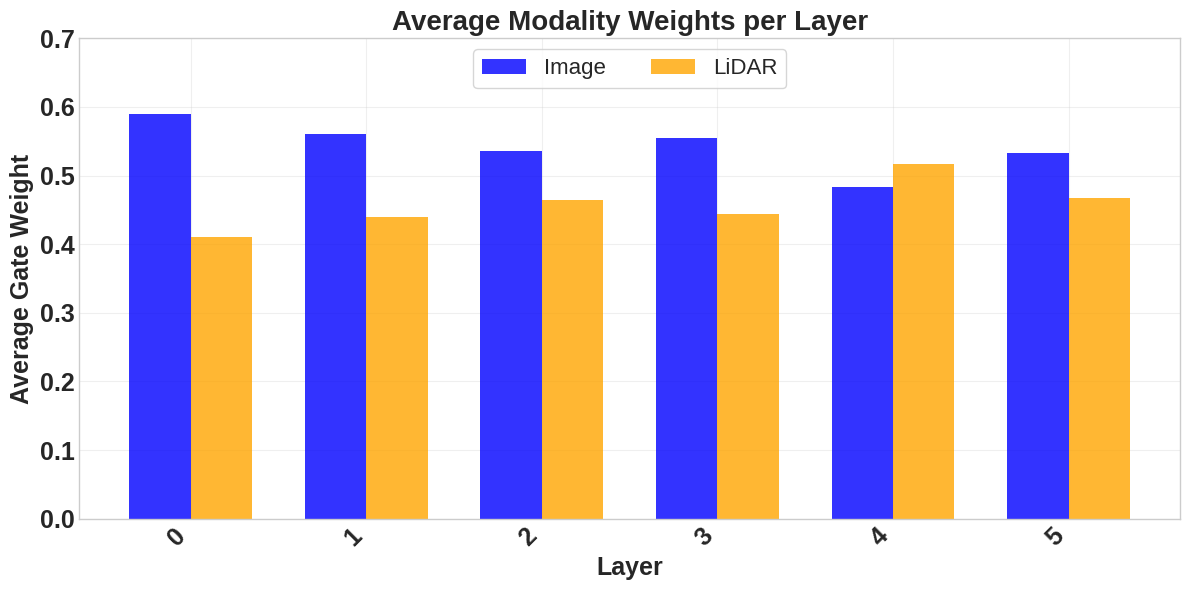}
\caption{Average weight of RGB and LiDAR modalities across all refinement layers of Li-ViP3D++ on testing data.}
\label{fig:lidar_modality_weight}
\end{figure}

This monotonic trend is consistent with an adaptive fusion behavior where the model strengthens geometric conditioning when reliable 3D evidence is available, which can directly reduce hallucinated detections. Importantly, the nonzero reliance even at very low point counts suggests that the model learns to exploit sparse LiDAR cues (e.g., a few returns at characteristic heights) without becoming brittle when LiDAR is weak.

Our analysis also showed the importance of continuous refinement of agent queries as per the DETR3D protocol. Each transformer decoder layer is able to utilize a different ratio of modality information for adjusting agent query representations as shown in Fig.~\ref{fig:lidar_modality_weight}.

\subsection{Robustness under reduced infrastructure and visual quality} 

The robustness experiments provide an important complement to the main comparison. The gains of Li-ViP3D++ are not confined to the default setting with HD map support and high-resolution RGB input, but remain visible when one or both of these sources of information are weakened. This suggests that the proposed fusion mechanism improves the intrinsic grounding of agent queries rather than merely exploiting a favorable input regime.

The results without HD map input are particularly informative. Although HD maps provide strong structural priors for motion forecasting, removing them does not eliminate the end-to-end advantage of Li-ViP3D++. The model continues to achieve the strongest EPA and the lowest false-positive tendency among the compared methods. This indicates that the enhanced query representation is able to recover useful scene context directly from the sensor inputs and from agent-level interaction cues, reducing reliance on externally provided map structure.

The reduced-resolution experiments point to a second form of robustness. For vision-heavy pipelines, downsampling RGB input would typically be expected to weaken both detection and forecasting performance. In contrast, Li-ViP3D++ maintains strong results and even improves on a few metrics (Miss Rate, Recall, mAP). A plausible interpretation is that the proposed multimodal fusion improves the balance between semantic visual cues and geometric LiDAR evidence, allowing the model to remain well grounded even when the visual branch contains less fine-grained detail.

From a deployment perspective, these findings are significant. HD maps may be unavailable, outdated, or costly to maintain, which limits the usability of models in new roads, construction zones, temporary road layouts, or low-cost deployments that cannot rely on continuously updated map infrastructure. However, removing map context does not make the model fully map-independent in a semantic sense. The map-free variant can still model agent-agent interactions and recover local scene cues from RGB and LiDAR, but it loses explicit lane topology, traffic-rule priors, and long-range road-structure information that could be important for forecasting rare maneuvers or resolving ambiguous intersections.

Image resolution is often limited by compute, sensing, or communication constraints. The fact that Li-ViP3D++ preserves its main advantages under both types of degradation suggests that the proposed fusion strategy is not only more effective in the nominal setting but also more resilient under practical operating conditions.

% TODO rewrite
\section{Conclusions and Future Work}
In this work, we introduced Query-Gated Deformable Fusion (QGDF), a query-space camera–LiDAR fusion mechanism that strengthens end-to-end perception and trajectory prediction. We apply QGDF to Li-ViP3D++, extending our prior Li-ViP3D architecture with differentiable LiDAR BEV sampling using learned per-query offsets and soft weighting, and query-adaptive modality gating that dynamically balances visual and LiDAR evidence. Together, these changes replace heuristic fusion and discrete top-$k$ selection with a fully differentiable, query-conditioned fusion process that is easier to optimize and more stable during training.

Through comprehensive evaluations on the nuScenes dataset, we demonstrate that Li-ViP3D++ significantly reduces displacement errors, increases robust prediction quality metrics such as EPA, and improves the detection quality measured by mAP, Precision, Recall and False Positive detection ratios. Our results indicate that query-based multimodal representations provide a robust interface for jointly optimizing perception and prediction.

On nuScenes, Li-ViP3D++ achieves consistent gains in both end-to-end behavior and detection quality. In particular, it reaches an EPA of 0.505 while improving detection performance to 0.616 mAP, alongside higher Precision/Recall and a substantially lower false-positive ratio of 0.069. Importantly, these robustness and calibration improvements do not come at a latency penalty. Li-ViP3D++ is slightly faster than the original Li-ViP3D baseline (139.82 ms vs. 145.91 ms), supporting its practical deployability. Because Li-ViP3D++ maintains strong performance under reduced RGB quality and without HD map input, it is suitable for practical settings where map infrastructure is unavailable, outdated, or costly to maintain. Nevertheless, the map-free setting should be viewed as a robust fallback rather than a complete replacement for semantic map context. Explicit lane topology and traffic-rule priors may become especially important in complex or ambiguous scenes, so the performance of the map-free variant in these cases should be evaluated more directly in future work. Even under this constrained setting, Li-ViP3D++ outperforms the original unimodal ViP3D and multimodal Li-ViP3D models while reducing inference latency by 36\% and 48\% respectively. 

Some future work still remains, namely that the strong performance of the proposed multimodal information sharing mechanism should also be evaluated to quantitatively determine the benefits it can bring under adverse conditions, such as lighting and weather variations. Overall efficiency of the model could likely also be improved by utilizing an input-dependent query initialization approach instead of the current learned distribution of agent queries standard in DETR3D-derived approaches. This could reduce the number of queries needed to reliably cover all agents present in a scene while increasing the overall efficiency of the model.

\bibliographystyle{IEEEtran}
\bibliography{IEEEexample}

% \begin{thebibliography}{00}
% \bibitem{b1} G. Eason, B. Noble, and I. N. Sneddon, ``On certain integrals of Lipschitz-Hankel type involving products of Bessel functions,'' Phil. Trans. Roy. Soc. London, vol. A247, pp. 529--551, April 1955.
% \bibitem{b2} J. Clerk Maxwell, A Treatise on Electricity and Magnetism, 3rd ed., vol. 2. Oxford: Clarendon, 1892, pp.68--73.
% \bibitem{b3} I. S. Jacobs and C. P. Bean, ``Fine particles, thin films and exchange anisotropy,'' in Magnetism, vol. III, G. T. Rado and H. Suhl, Eds. New York: Academic, 1963, pp. 271--350.
% \bibitem{b4} K. Elissa, ``Title of paper if known,'' unpublished.
% \bibitem{b5} R. Nicole, ``Title of paper with only first word capitalized,'' J. Name Stand. Abbrev., in press.
% \bibitem{b6} Y. Yorozu, M. Hirano, K. Oka, and Y. Tagawa, ``Electron spectroscopy studies on magneto-optical media and plastic substrate interface,'' IEEE Transl. J. Magn. Japan, vol. 2, pp. 740--741, August 1987 [Digests 9th Annual Conf. Magnetics Japan, p. 301, 1982].
% \bibitem{b7} M. Young, The Technical Writer's Handbook. Mill Valley, CA: University Science, 1989.
% \end{thebibliography}
% \vspace{12pt}
\end{document}